\documentclass[10pt,twocolumn,letterpaper]{article}

\usepackage{wacv}
\usepackage{times}
\usepackage{epsfig}
\usepackage{graphicx}
\usepackage{amsmath}
\usepackage{amssymb}

\usepackage{color}
\usepackage{amsfonts}
\usepackage{multirow}
\usepackage{paralist}
\usepackage{rotating}
\usepackage{subfigure}
\usepackage{bm,mathrsfs}
\usepackage{booktabs}
\usepackage{tabu}

\wacvfinalcopy 

\def\wacvPaperID{313} 

\ifwacvfinal\fi
\begin{document}
\newcommand{\addressnet}{AddressNet}
\newcommand{\fastaddressnet}{Enhanced AddressNet}
\newcommand{\addressbased}{Address-based module}
\newcommand{\addressenhanced}{Address-enhanced module}
\newcommand{\addmobile}{Address-MobileNet}
\newcommand{\WBC}[1]{{\color{red}{[}WBC: #1{]}}}
\newcommand{\liuxg}[1]{{\color{blue}{[}liuxg: #1{]}}}
\newcommand{\ZHS}[1]{{\color{green}{[}ZHS: #1{]}}}
\newcommand{\SH}[1]{{\color{red}{[}SH: #1{]}}}
\title{Shift-based Primitives for Efficient Convolutional Neural Networks}

\author{Huasong Zhong\thanks{indicates equal contribution}$^{*2}$ \hspace{1cm} Xianggen Liu$^{*2}$ \hspace{1cm} Yihui He$^{*1}$ \hspace{1cm} Yuchun Ma$^2$\\
$^1$Carnegie Mellon University \hspace{1cm} $^2$Tsinghua University\\
{\tt\small he2@andrew.cmu.edu  \{zhonghs16,liuxg16,myc\}@mails.tsinghua.edu.cn}
}

\maketitle
\ifwacvfinal\thispagestyle{empty}\fi

\begin{abstract}
We propose a collection of three shift-based primitives for building  efficient compact CNN-based networks.
These three primitives (\textbf{channel shift, address shift, shortcut shift}) can reduce the inference time on GPU while maintains the prediction accuracy. These shift-based primitives only moves the pointer but avoids memory copy, thus very fast. For example, the channel shift operation is 12.7$\times$ faster compared to channel shuffle in ShuffleNet but achieves the same accuracy. The address shift and channel shift can be merged into the point-wise group convolution and invokes only a single kernel call, taking little time to perform spatial convolution and channel shift. Shortcut shift requires no time to realize residual connection through allocating space in advance. We blend these shift-based primitives with point-wise group convolution and built two inference-efficient CNN architectures named \addressnet\ and \fastaddressnet. Experiments on CIFAR100 and ImageNet datasets show that our models are faster and achieve comparable or better accuracy. 
\end{abstract}
\section{Introduction} 

Convolutional neural networks (CNNs) have been firmly established as the prevalent methods in image understanding problems such as image classification,  image caption, and object detection~\cite{imagenet,resnet,rcnn,fasterrcnn,fcn}. The high accuracy is at the cost of increased computation time and memory usage. Real-time processing is vital in some applications such as self-driving cars and speech recognition, where low latency, small storage, and an appropriate accuracy are required~\cite{mobilenet,shufflenet,ma2018vehicle}. Thus producing fast and energy efficient CNNs are very well motivated. 


There are a number of recent efforts aimed at reducing CNN model size and computational requirements while retaining accuracy. For example, MobileNets~\cite{mobilenet} propose a family of lightweight neural networks based on depthwise separable convolution. ShuffleNet~\cite{shufflenet} utilizes pointwise group convolution and channel shuffle to reduce parameters and FLOPs. To further decrease parameters, ShiftNet~\cite{shiftnet} adopts the shift operation on a feature map as an alternative to spatial convolution.  Unfortunately, smaller parameter size or number of FLOPs do not always lead to direct reduction of actual inference time, since many core operations introduced by these state-of-the-art compact architectures are not efficiently implemented for GPU-based machines. For instance in MobileNet, depthwise separable convolutions \textbf{only consume 3\% of the total FLOPs and 1\% of the parameters, but they constitute 20\% of the total inference time}. Channel shuffle and shortcut connections in ShuffleNet do not require any FLOPs or parameters; however, these operations still constitute 30\% of the total inference time. Similarly, in ShiftNet the feature map shift is \textbf{parameter-free and FLOP-free, but it occupies 25\% of the total inference time}. Although MobileNet and ShuffleNet have roughly the same FLOPs, the latter requires two times more inference time. More details are shown in Table~\ref{tab:statistics}. Therefore, in practice, \textbf{neither reducing parameters nor FLOPs ensures a reduction in inference time}.
\begin{figure}[t]
	\centering
	\includegraphics[width=0.98\linewidth]{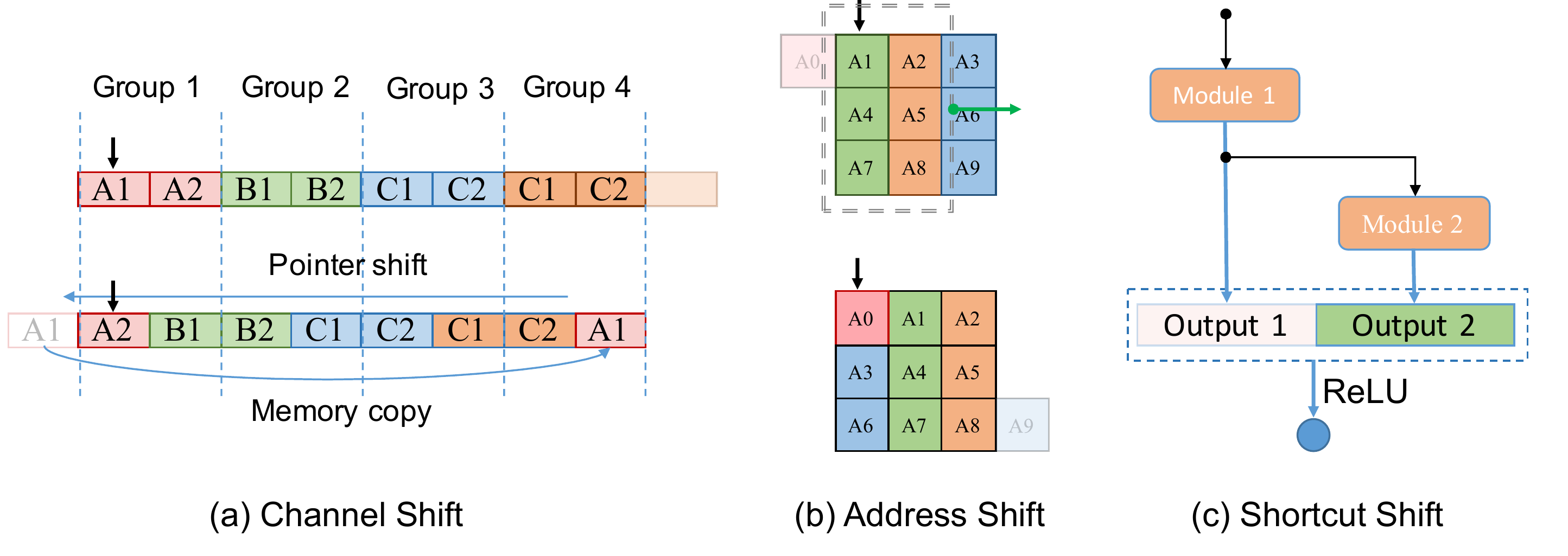}
	\caption{Three efficient shift primitives for efficient neural network architecture design}\label{f:three_operations}
\end{figure}
Based on the above concerns, we propose a collection of three shift primitives (Fig.~\ref{f:three_operations}) for CNN-based compact architectures to reduce parameters, FLOPs and inference time simultaneously for GPU-based machines: 1) the channel shift  that acts as a faster alternative to the channel shuffle operation. 2) the address shift that efficiently collects spatial information at no cost of actual inference time.  3) the shortcut shift provides a fast channel concatenation to realize residual connection through allocating continuous memory space in advance, but does not consume inference time. This collection of primitives mainly involves moving pointers in continuous memory space to minimize actual memory copy and completely avoid floating-point operations, which leads to actual speedup.
We combined these collection of shift primitives together with point-wise group convolution to build two compact architectures named \addressnet\ and \fastaddressnet\ respectively.  Experiments on CIFAR100 and ImageNet datasets demonstrate that our models can achieve equal or superior accuracy with less inference time.

\section{Related Work}
\label{s:related}
Deep convolutional neural networks provide the best results on many computer vision tasks~\cite{imagecaption,fcn}. However, deep neural networks are not always computationally efficient and there appears to be redundant computation in most models. The desire to deploy accurate deep neural networks in low latency applications natural motivates the search for methods to decrease model size and operations (parameters, FLOPs), and, more generally, overall inference time. Recently \cite{smallnn} surveyed current approaches to designing small, energy efficient deep neural nets.  Here we take another approach and note that  these approaches can be categorized into either compressing pre-trained models or training small models directly. 

\subsection{Compressing Neural Networks} 

We outline four types of widely used methods for model compression. First, pruning is an approach that reduces redundant parameters that are insensitive to the accuracy~\cite{channelpruning,amc,thinet,networktrimming,deepcompression}. Second, low-rank factorization, which estimates the informative parameters by matrix decomposition, has been used in~\cite{lowrank,linearstructure,lowrankregularization,channeldecomposition}. 
Thirdly, quantization and binarization~\cite{vectorquantization,quantizedconvolution,xnornet} can reduce the number of bits to represent each weight. Lastly, knowledge distillation~\cite{bayesian,knowledgedistill} is a method that trains a small neural network by distilling knowledge of a large model. We adopt a different approach from these because we design compact network directly instead of compressing a pretrained network.

\subsection{Designing compact layers and networks}  There has been increasing interest in building efficient and small models, \eg\ \cite{flattened,quantizedconvolution,xnornet,lightweight}.  In ResNet~\cite{resnet}, the bottleneck structure has been proposed to decrease the channel numbers before and after a 3 $\times$ 3 convolution. ResNeXt~\cite{resnext} introduces a multi-branch and homogeneous architecture to decrease FLOPs and improve accuracy. The fire module is introduced in SqueezeNet~\cite{squeezenet} where a fraction of 3 $\times$ 3 convolutions is replaced by 1$\times$1 convolutions to reduce parameters. GoogLeNet~\cite{googlenet} is a well-designed network with a complex structure to reduce computation and increase accuracy. More generally, in order to reduce the amount of parameters and FLOPs, the following operations may be used:

\textbf{Depthwise Separable Convolution.} The initial work on depthwise separable convolution is reported in Sires's thesis which is inspired by prior research from Sifre and Mallat on transformation-invariant scattering~\cite{scattering}. Later, Inception V1 and Inception V2 used it as the first layer. After that, it was adopted by MobileNet to design network architecture for mobile devices. The depthwise separable convolution applies a single filter to each input channel. 
While this operation can reduce computation in theory, in practice it is hard to implement a depthwise separable convolution layer efficiently because a fragmented set of memory footprints are required. Even though there are few FLOPs in a layer, it is still very expensive for inference time as Table~\ref{tab:statistics} shows, and this drawback is also mentioned in~\cite{xception,shufflenet}. 
    
\textbf{Feature Map Shift.} ShiftNet~\cite{shiftnet} presents a parameter-free, FLOP-free shift operation as an alternative to spatial convolution. It can be viewed as a special case of depthwise separable convolution that results from assigning one of the values in each $n \times n$ kernels to be 1 and the rest to be 0. It can not be implemented efficiently either by depthwise separable convolution. The address shift operation, based on pointer shifting, varies from ShiftNet obviously. To avoid confusion, in this paper, we use the term \textit{feature map shift} to refer specifically to the method proposed by ShiftNet.

\textbf{Pointwise Group Convolution and Channel shuffle.} ShuffleNet~\cite{shufflenet} adopts pointwise group convolutions to reduce parameters and FLOPs, but it  also brings the side effect of blocking information flow between group convolutions. Channel shuffle is proposed to address this problem. This is also difficult to implement efficiently since a channel shuffle will move the entire set of channels into another memory space.
    
\textbf{Residual Connection.}  Residual connections are introduced by He et al~\cite{resnet} to
enable	very smooth	forward/backward propagation. There are two categories of shortcut connection operations including identity mapping and channel concatenation. Both of them are used in ResNet and ShuffleNet. DenseNet~\cite{densenet} concatenates all previous output of layers before the activation function. 

\begin{table*}[t]
		\begin{center}
			\begin{tabular}{ccccccc}
			\hline
			Network & FLOPs & Params & Operation & Parameter (\%) & FLOPs (\%) & Time (\%) \\ \hline
			ShuffleNet & 524M & 5.6M & identify map & 0 & 0 &6\%\\ 
			ShuffleNet & 524M & 5.6M & channel shuffle & 0 & 0 &24\%\\ 
			MobileNet & 569M & 4.2M & depthwise conv & 3\% & 1\% &20\%\\
			ShiftNet & 1.4G & 4.1M & feature map shift & 0 & 0 &25\% \\
			\hline
			\end{tabular}
		\end{center}
        			\caption{Comparison of number of FLOPs, number of parameters and inference time in different operations and models}
			\label{tab:statistics}
\end{table*}

\section{Approach} 
In this section, we introduce the three shift primitives which we find sufficient for producing fast CNN models. 
We then fuse these shift primitives together with pointwise group convolution to create two efficient modules and from these we create novel deep neural networks.

\subsection{Channel Shift} 
\label{s:channel-shift}
\begin{figure}[!t]
	\centering
	\includegraphics[width=\linewidth]{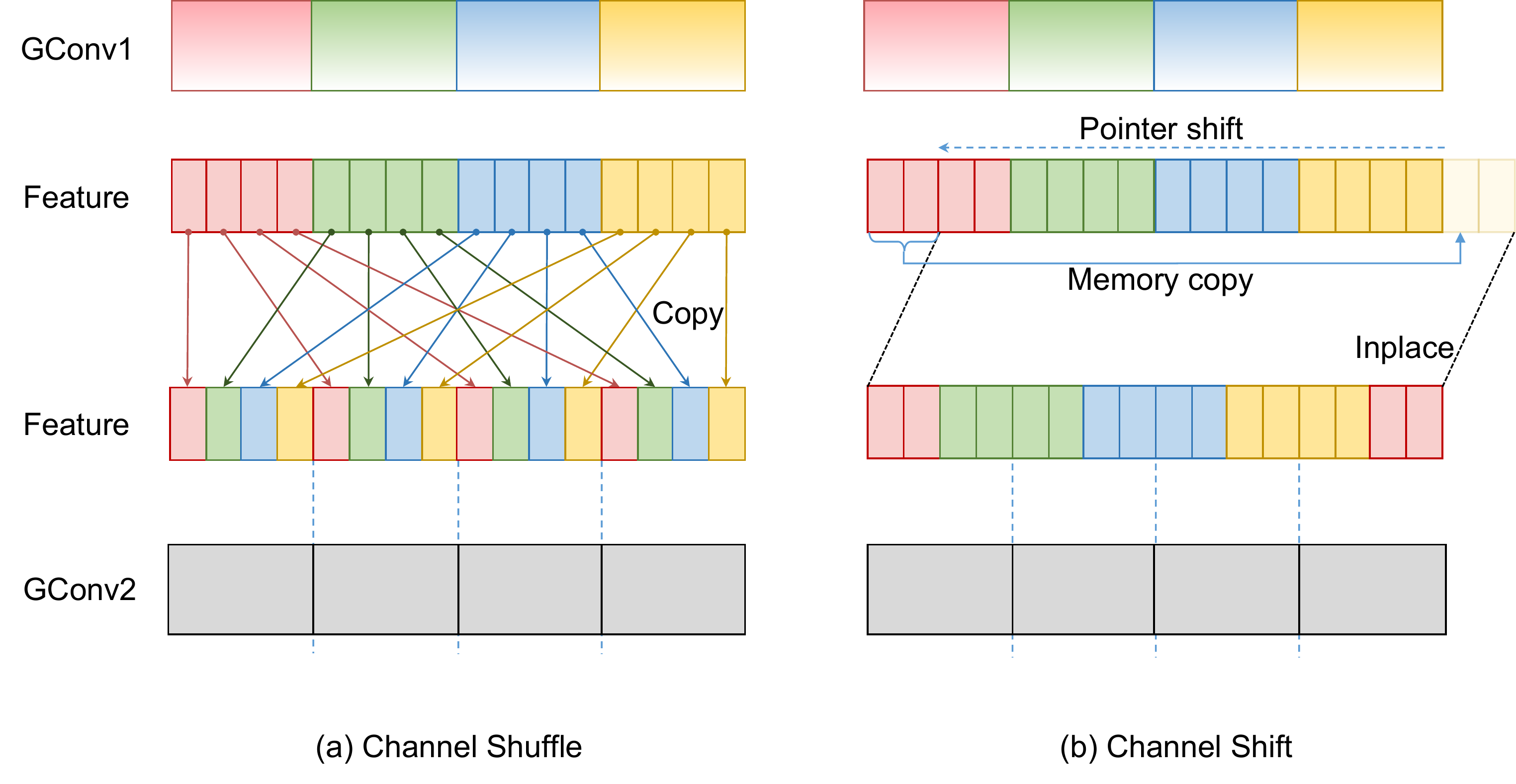}
	\caption{The computation diagrams of channel-shuffle and channel-shift, both with two stacked convolution layers. a): channel shuffle layer with the same number of groups. Input channels are fully mixed when GConv2 takes data from different groups after GConv1, and each colored arrow denotes copying data one time. b): channel shift layer, where the channels are shifted circularly along a predefined direction and there the process at most spends two units to copy data, thus 8$\times$ less memory movement (in this example) than channel shuffle}\label{f:channel-shift}
\end{figure}

Group convolution can reduce the computational complexity of ordinary convolution. 
For example, AlexNet~\cite{alexnet} used group convolutions to divide the feature map across two GPUs. More generally, a group convolution with group number $G$ reduces the FLOPs and parameter size by a factor of $G$.  However, stacking group convolution layers together can block the information from flowing among groups and reduce accuracy. To mitigate this, the channel shuffle operation in ShuffleNet~\cite{shufflenet} is adopted to fuse features among different groups. As illustrated in the left panel of Figure~\ref{f:channel-shift}, channel shuffle is time-consuming since it requires moving feature maps to another memory space. Note that moving data is much more expensive in terms of latency and energy consumption compared with floating point operations~\cite{eie}. In contrast, shifting the pointer, or the physical address to load data, is free. Therefore, we propose the channel shift primitive to utilize pointer shift and minimize actual data movement to reduce the time and energy. 

The channel shift primitive blends the information in adjacent channels by shifting all the channels along a certain direction. Taking four-group convolution for example, as shown in Fig.~\ref{f:channel-shift} (b), a predefined storage space (\ie\ the two yellow grids) is allocated at the end of the feature map, and half of the first group is moved there. Then, the pointer is shifted backwards by two grids (half of a group). Note that this primitive does not involve any data movement. When the second group convolution ($GConv2$) fetches data from the shifted address, each group in $GConv2$ can get data from two groups of input feature map. Moving data circularly, a single layer of channel shift fuses information between adjacent groups and stacking multilayers can fuse more. This is not equivalent to channel shuffle method, but our experiment show that channel shift lead to similar accuracy. Furthermore, compared to channel shuffle whose mapping is more complex and requires much more actual data movement, channel shift is needs 8$\times$ less data movement in this case because it only needs to copy 2 units of data while channel shuffle needs to copy 16 units, as illustrated in Figure~\ref{f:channel-shift}.

\subsection{Address Shift} 
Modern convolutional neural networks usually consist of convolution layers with different kernel sizes and channels. Nevertheless, as 90\% or more  time is spent in convolutions, improving the process of convolutions is attractive.
Fortunately,  feature map shift can provide the equivalent function of spatial convolutions with zero FLOPs and no parameters. Despite all this, feature map shift in ShiftNet~\cite{shiftnet} still consumes inference time heavily. In order to solve this problem, we propose the address shift primitive and this will be detailed in the following.

\begin{figure}[!t]
	\centering
	\includegraphics[width=0.98\linewidth]{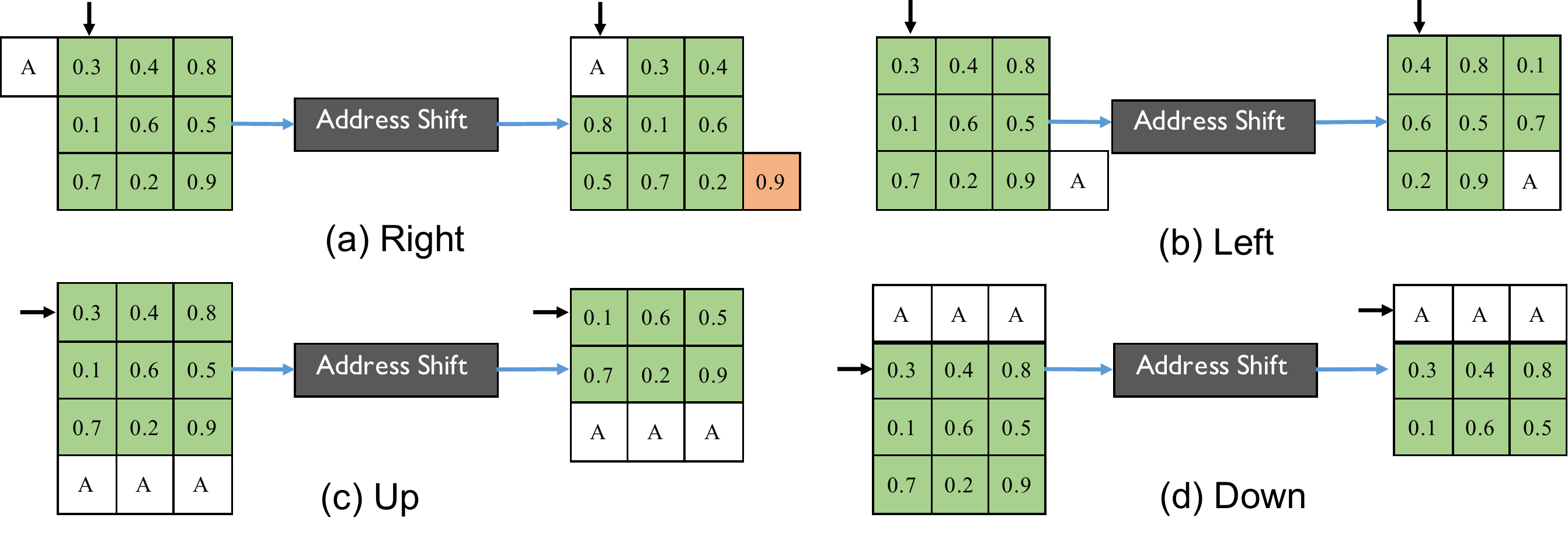}
	\caption{The implementation of address shift in four directions, where A stands for values from adjacent feature map, and the black arrow denotes the address of feature map pointer}\label{f:address-shift}
\end{figure}
Figure~\ref{f:address-shift} presents  address shift primitives in four different 
directions. Taking right shift for example, as illustrated in figure
\ref{f:address-shift} (a), 
a shift pointer is offset by 1 unit ahead of the starting address of the feature map and is pointed to address A. Then fetching the tensor continuously in the memory space starting from address is equivalent to shifting the entire tensor to the right by one grid. 
Similarly we can define the other three shift operations of different directions(left, up and down): moving the pointer one unit forward is equivalent to left shift, and skipping (backtracking) a row is equivalent to shift down (up), as illustrated in the rest parts of Figure~\ref{f:address-shift}. Formally, the above process can be abstracted as the following formula:
\begin{align}
	\bm x_r &=tensor(p_{\bm x}-s_d)
\end{align}
where function $tensor(p)$ denotes a read operation to get a tensor at pointer $p$, $s_d$ denotes the offset for shift direction $d$. More specifically, $s_{right} = 1$, $s_{left} = -1$, $s_{up} = -stride$, $s_{down} = stride$.

\begin{figure}[!t]
	\centering
	\includegraphics[width=\linewidth]{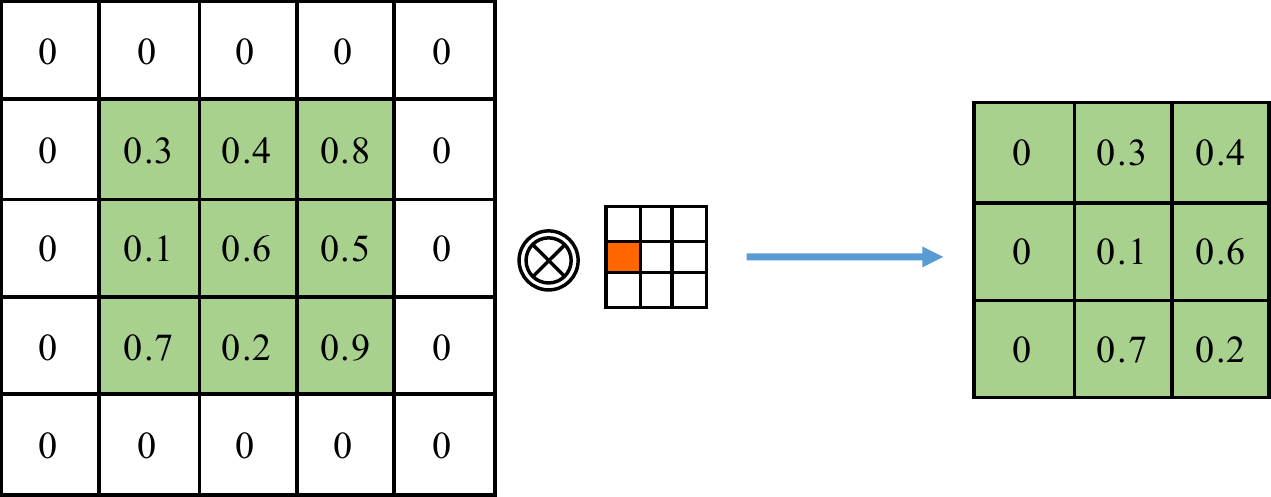}
	\caption{Implementation of feature map right shift by depthwise convolution, proposed by \cite{shiftnet}, where the result is similar to address shift operation in Figure~\ref{f:address-shift}(a)}\label{f:shiftnet}
\end{figure}

The proposed operation is functionally similar to the shift mechanism in ShiftNet~\cite{shiftnet}, which is implemented by a predefined spatial convolution where the kernel only contains one non-zero value to indicate shift direction, as shown in Figure~\ref{f:shiftnet}. The feature map shift can be seen as a special case of depthwise separable convolution. Comparing Figure~\ref{f:shiftnet} and Figure~\ref{f:address-shift}(a), we can see that one difference is with the boundary, where the feature map shift will lead to 0-padded boundaries, but address shift will have none-zero boundaries. In our experiment however, we found this nuance does not have any noticeable effect to the network's accuracy. 

Based on the four basic address shift operations (up, down, left, right), we can compose arbitrary shift patterns (such as up-left). In practice, the cost of this operation is still expensive in inference, since the number of possible shift directions grows quadratically with respect to kernel size (3x3 kernel: 9 possible directions; 5x5 kernel: 25 possible directions), as explained in \cite{shiftnet}. Inspired by convolution decomposition in \cite{rethinking}, 3$\times$3 convolution can be divided into 1$\times$3 and 3$\times$1 convolution. Similarly, top-left shift direction can be decomposed into top and left shift. With the increasing of channel number, CNNs equipped with address shift operation can fuse all information from each direction. Thus we can just use \textbf{four fundamental shift directions} to represent other directions to simplify the network architecture. 

\subsection{Shortcut Shift}
With the success of ResNet, shortcut connections have become common in  deeper network architectures. 
Both addition and concatenation are effective implementation choices, and concatenation performs better according to the analysis in SparseNet~\cite{sparsenet}. Shortcut connections integrate the lower and detailed information with the higher representation and offers a shorter path for back propagation and channel concatenation does not require any computation, and can therefore lead to faster inference speed than addition. 

We propose a further optimization for channel concatenation: a fixed-size space is allocated in advance which places the output of current layer right after the output of the last layer. In other words, our approach can make the output of two layers located in a pre-allocated continuous storage space so that no copy or computation time is spent on channel concatenation. Considering the starting pointer of  current output  is shifted to the end of last output, we name it ``shortcut shift'', in accordance with our naming style.  This optimization can be better leveraged on DenseNet~\cite{densenet} which heavily relies on channel concatenation.

\subsection{Address-based module}
\label{s:architeture}
\begin{figure}[t]
	\centering
	\includegraphics[width=\linewidth]{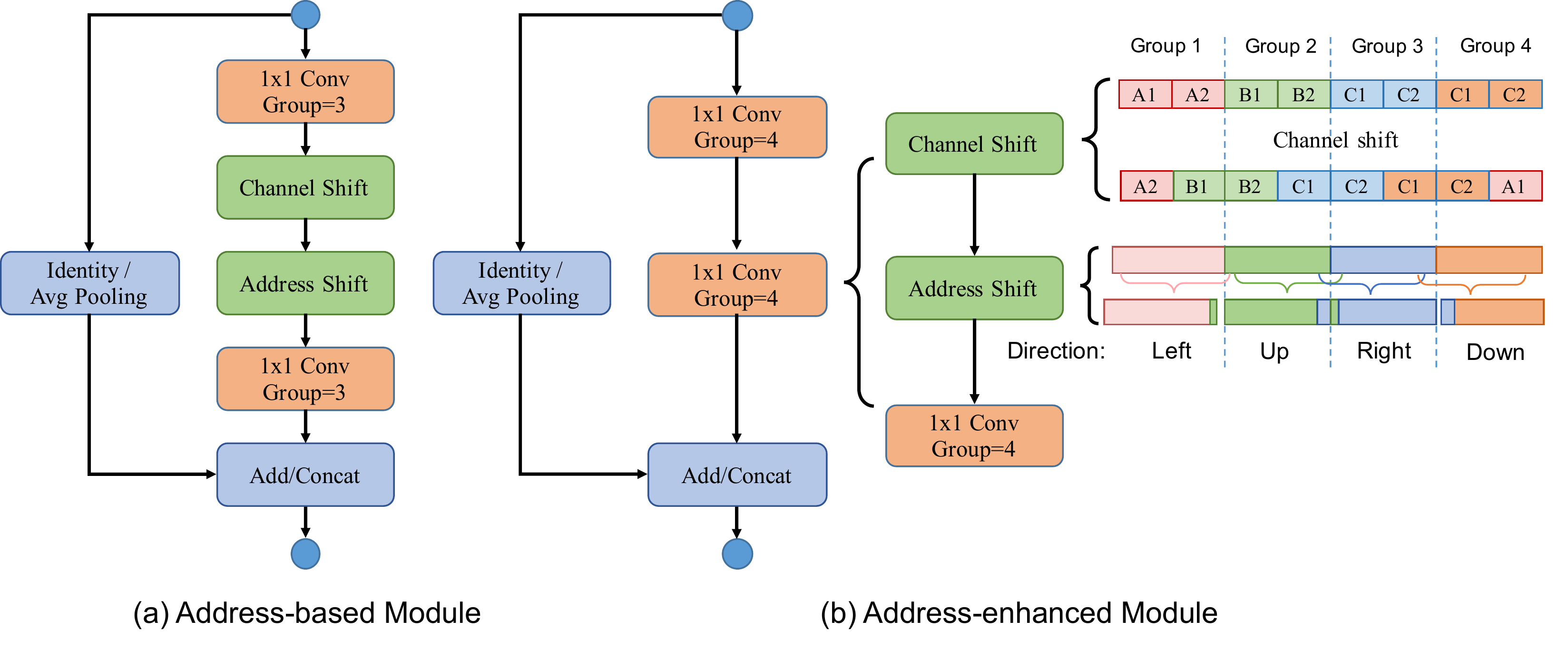}
	\caption{\addressbased\  and \addressenhanced. a): the unit with pointwise group convolution (GConv), channel shift and address shift, where only the GConv consumes parameters. b): similar to the left one, but different in with regard to the second GConv operation in which we embed channel shift and address shift. This is expected to significantly reduce computation. }\label{f:architeture}
\end{figure}

Based on the  address shift and channel shift operations described above, we build an \addressbased\ as a collection of layers in the manner of the bottleneck module in ResNet~\cite{resnet}. As shown in Figure~\ref{f:architeture}(a), we use the pointwise group convolution layer in the beginning. Then the channel shift layer exchanges information among channel groups. Next, the address shift layer mixes spatial information. In the address shift module, we divide the channels into 3 groups, and within each group, the address shift operation moves the data towards four directions of \{up, down, left, right\}. Finally, we perform another pointwise group convolution to fuse information and match the output channel dimension. Following ShuffleNet~\cite{shufflenet}, we use an additive residual connection if the size of feature map maintains. Otherwise, we use average pooling and concatenation. The first group convolution is followed by batch normalization and non-linear activation function(ReLU) while the second is just followed with batch normalization. 

We find that both channel shift and address shift operation can be embedded into the second pointwise group convolution to speed up. Thus we design an \textit{enhanced} group convolution to realize channel shift, address shift and group convolution together. For simplicity,  we set the number of groups to four, each group corresponds to one direction for address shift operation. Also, taking address offset into consideration, as depicted in Figure~\ref{f:architeture} (b), we suggest that shifting left and up (forward offset) are performed in the first two group, and shifting right and down (backward offset) for the last two group. In practice, this design is conducive to implement without memory overflow or extra memory overhead.  Thus, we call the \addressbased\ equipped with the above-sophisticated design as \addressenhanced.

\section{Experiments}
As described above, we present a collection of shift operations including channel shift, address shift and shortcut shift, to reduce parameters, FLOPs, and more importantly, actual inference time, while retaining accuracy. In this section, we first build variants of our compact networks, called \addressnet\ and \fastaddressnet,  based on the above two basic modules. We then investigate their basic functions by comparing with other primary operations on CIFAR100 dataset. We then compare the two architectures with ShiftResNet~\cite{shiftnet}. Finally, we use the proposed approaches to modify MobileNet and evaluate it, as well as our own models, on the ImageNet dataset. 

In \addressnet\ and \fastaddressnet, a 3x3 convolution is first applied with 36 filters and 16 filters. Then three blocks are stacked with \addressbased\ and \addressenhanced\ on the feature map sizes \{32, 16, 8\} respectively. The subsampling is performed in the second pointwise group convolution with a stride of 2. 
An identity map is performed when the adjacent units contain the same feature map size. An average pooling is used to match the shape and double the output channels. The output channel numbers of blocks are \{48, 60, 96\} and \{48, 96, 192\} respectively. The network ends with a global average pooling followed by a 100-way fully-connected layer and softmax classifier. Each block contains 3 modules in \addressnet-20 and \fastaddressnet-20, 5 modules for \addressnet-32 and  \fastaddressnet-32, and 7 modules for \addressnet-44 and \fastaddressnet-44.

\subsection{Implementation details}  
To improve our experimental results, hyperparameters are fine-tuned with a coarse grid search. Following ShiftResNet, ``expansion rate"($\varepsilon$) is used to scale the number of channels in the intermediate layers.  For simplicity, we set the expansion rate to 3 for most experiments on CIFAR100 dataset and the $\varepsilon$ for ImageNet is shown in Table~\ref{table:addressnetArch}.  On the CIFAR100 dataset, the input image is 32$\times$32.  We use a stochastc gradient-descent (SGD) optimizer with mini-batch 128 on 4 GPUs (\ie\ 32 per GPU). The weight decay is 0.0005 with momentum 0.9 and training the network for 300 epochs. The learning rate starts with 0.1 and decreases by a factor of 10 after 32k and 48k iterations and a factor of 2 from 64k to 128k in every interval 16k iterations. We adopt the weight initialization of \cite{msra}. While on the ImageNet dataset, the input image is randomly cropped from a 256$\times$ 256 image to 224$\times$ 224. We start from a learning rate of 0.1 with divided by 10 for every 30 epochs and weight decay of 0.0001. We evaluate the error on the single 224x224 center crop from an image whose shorter side is 256. During testing, we remove all batch normalization (BN) layers because they can be fused into a convolution layer in advance. We implement our network in Caffe framework and test inference time on a single GeForce GTX 1080 GPU for two epochs with a batch size of 1 and report the average performance.

\subsection{Shift-based Primitives vs. baselines}
In the following experiments, we will first study the equivalence between our proposed operations and their baselines on CIFAR100 dataset. 

\subsubsection{Channel shift vs. Channel shuffle}
Channel fusion is especially critical to the performance of small networks. Channel shift and channel shuffle promote the information fusion to varying degrees. 
To compare the performance of channel shift and channel shuffle, we replace all channel shifts in AddressNet-32 with channel shuffle and the results is shown in Table~\ref{table:channel-perf}. The two models achieved the same accuracy, but AddressNet-32 equipped with channel shift is much faster than channel shuffle, with 1.4 times faster in total time and 12.7 times faster in operation time(operation time denotes the average time that the operation consumes). It is quite interesting that the acceleration is larger than theoretical estimation that is described in Section~\ref{s:channel-shift}. This shows advantages of embedding the shift operations: as introduced in Section~\ref{s:architeture}, embedding the channel shuffle into group convolutions saves inference time significantly.

\begin{table*}[t]
\begin{center}
		\begin{tabular}{cccccc}
			\hline\noalign{\smallskip}
			Model & Params & FLOPs & Top1 Accuracy & Total Time & Operation Time \\
			\noalign{\smallskip}
			\hline
			\noalign{\smallskip}
			\addressnet-32-shift & 0.14M & 37.3M & 71.96\% & \textbf{4.8} & \textbf{0.15}\\
			\addressnet-32-shuffle & 0.14M & 37.3M & 71.97\%  & 6.5 &  1.9\\
			\hline
            Speed Up & - & - & -  & 1.4$\times$ & 12.7$\times$ \\
            \hline
		\end{tabular}
\end{center}
		\caption{Comparison between Channel shift and Channel shuffle operations conditioned on the same size of parameters and FLOPs. The following two models conform to the architecture of AddressNet-32 and just differ in channel transformation. The total time(ms) means average runtime of the model and operation time(ms) denotes the average time that the operation consumes}
		\label{table:channel-perf}
\end{table*}

\subsubsection{Address Shift vs. Feature map Shift} 
Based on the analysis in the previous section, we argue that some shift directions are redundant given the four basic directions. 
To validate this, we first replace nine shift directions (kernel size of 3) on feature map with four fundamental shift directions in ShiftResNet network architecture~\cite{shiftnet}. Secondly, to compare performance between different shift operations, we replace feature map shift with our address shift in the same network architectures to build AddressResNet. The result is displayed in Table~\ref{table:headings}. Firstly, we notice that both two combinations of directions achieve nearly the same performance, which demonstrates the shift operation based on four basic directions is adequate for shifting feature map. Secondly, we find both address shift and feature map shift can achieve nearly the same performance. Thirdly, regarding the inference time, the model that uses address shift achieved nontrivial speedup. Thus, this experiment validates our expectation successfully, and in the following experiments, all the models will use address shift with four directions. 
 
 \begin{table*}[t]
 \begin{center}
		\begin{tabular}{cccccc}
			\hline\noalign{\smallskip}
			Model & Params & FLOPs & Top1 Accuracy & Total time & Operation Time \\
			\noalign{\smallskip}
			\hline
			\noalign{\smallskip}
			ShiftResNet-four & 145k & 22.9M & 71.80\% & 3.3 & 0.4 \\
			ShiftResNet-nine & 145k & 22.9M & 71.86\% & - & -\\
			AddressResNet-four & 145k & 22.9M & \textbf{71.94}\% & 3 & 0.3 \\
			AddressResNet-nine & 145k & 22.9M & 71.91\% & - & -\\
			\hline
            Speed Up & - & - & -  & 1.1$\times$ & 1.3$\times$ \\
            \hline
		\end{tabular}
\end{center}
		\caption{Performance of address shift and feature map shift operations following the above comparison method. The four models conform to the architecture of ShiftResNet~\cite{shiftnet} and just differ in feature map transformation and number of directions}
		\label{table:headings}
\end{table*}

\subsubsection{Shortcut shift vs. Identity map} The purpose of the shortcut shift operation is to achieve shortcut connections for free. We replace all the identity map layers in Fast \addressnet-20 with shortcut shift operation to build \fastaddressnet-20-concat.  To prevent over-accumulation of channels in deep layers, we reduce the output channels of current layer to match the output channel of previous residual function and set the expansion rate to four in all layers. This gives similar FLOPs and parameters. The result is shown in Table~\ref{table:shortcut-compare}.  We notice that they can achieve almost equivalent performance with the similar parameters and FLOPs. However, our shortcut shift does not need to spend any inference time obviously(so we ignore the comparison of inference time).

\begin{table}[t]
\begin{center}
		\begin{tabular}{cccc}
			\hline\noalign{\smallskip}
			Model & Params & FLOPs & Top-1\\
			\noalign{\smallskip}
			\hline
			\noalign{\smallskip}
			\fastaddressnet-20 & 0.18M & 27M & 69.45\% \\
			\begin{tabular}[c]{@{}c@{}}\fastaddressnet-20\\ (concat)\end{tabular} & 0.19M & 29M & 69.71\% \\
			\hline
		\end{tabular}
\end{center}    
\caption{Performance of two shortcut connection implementation, following the above comparison method}
\label{table:shortcut-compare}
\end{table}
\subsubsection{Address shift vs. Depthwise separable convolution}
When the outputs are used in a spatial aggregation context, there is strong correlation between adjacent units which results in much less loss of information during dimension reduction. 
Thus, depthwise separable convolutions are critical for extraction of spatial features. Address shift naturally derives spatial features in different views, so we integrate address shift into a CNN-style network to test its utility and expressive ability. To evaluate the performance fairly, we choose depthwise separable convolution to design MobileNet-32 whose number of layers is the same as \fastaddressnet-32. We follow most of the hyper-parameters in \fastaddressnet-32 with one exception: the output channels of first convolution and three blocks are 48 and \{56, 112, 224\} respectively in MobileNet-32. In table~\ref{table:shift-compare}, it shows that \fastaddressnet-32 can behave better with similar scales of parameters and FLOPs. It indicates that our address shift is equal or superior to depthwise separable convolutions in its expressive ability and capacity to extract the feature of images. 

\begin{table}[t]
\begin{center}
\begin{tabular}{cccc}
\hline\noalign{\smallskip}
Model & Params & FLOPs & Top-1  \\
\noalign{\smallskip}
\hline
\noalign{\smallskip}
\fastaddressnet-32 & 328k & 48M & \textbf{73.55}\%  \\
MobileNet-32 & 337k & 50M & 71.62\%  \\
\hline
\end{tabular}
\end{center}
\caption{Comparison in expressive ability(accuracy) between address shift and depthwise separable convolution conditioned on the same size of parameters and FLOPs. The following two models conform to the architecture of \fastaddressnet-32 and just differ in feature map transformation}
\label{table:shift-compare}
\end{table}

\subsection{Performance on CIFAR100}
\begin{figure*}[t] 
  \subfigure{  
  \begin{minipage}{5cm}
    \centering   
    \includegraphics[scale=0.22]{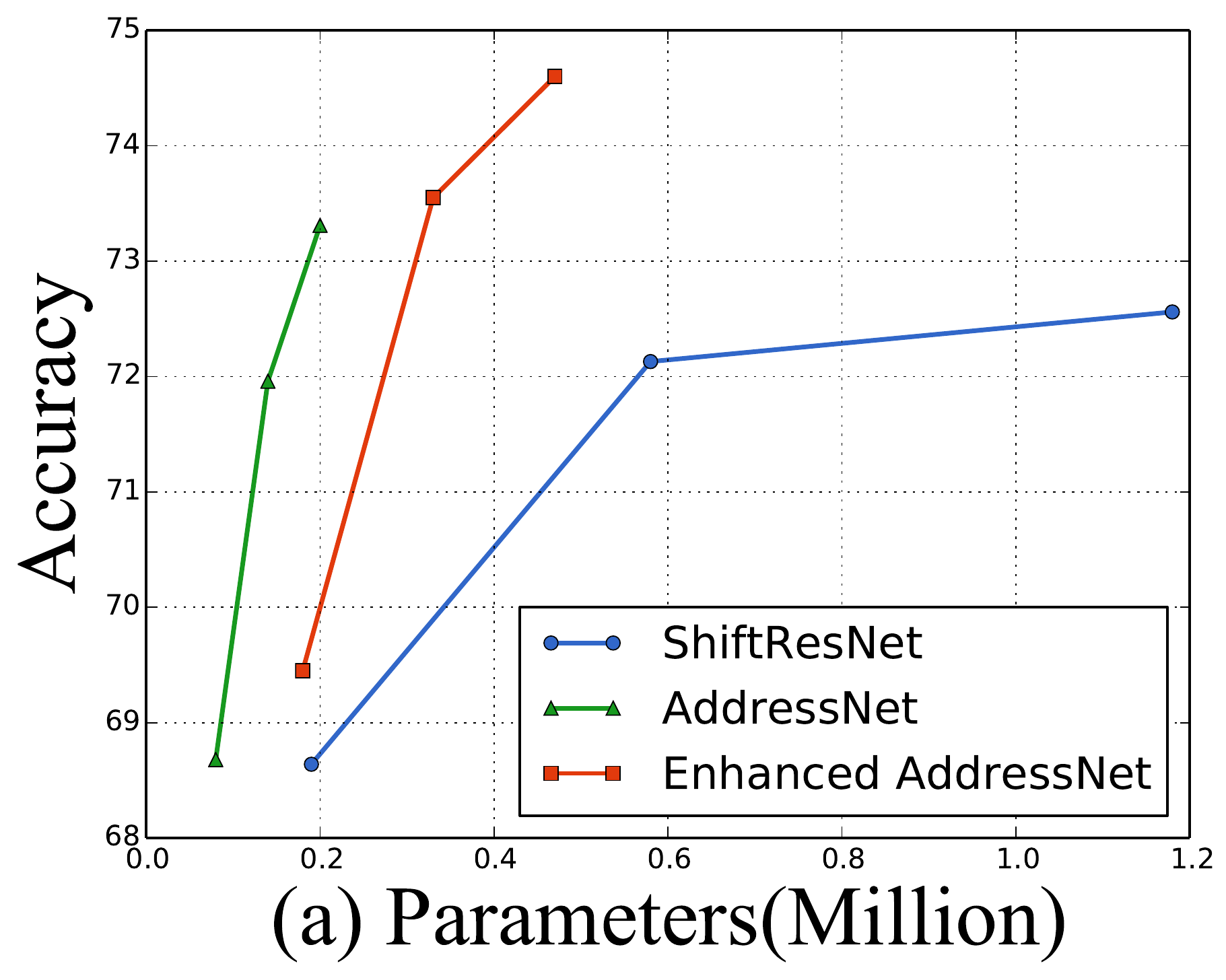}               
   \end{minipage}
   }
   \quad \quad
  \subfigure{  
    \begin{minipage}{5cm}
      \centering    
      \includegraphics[scale=0.22]{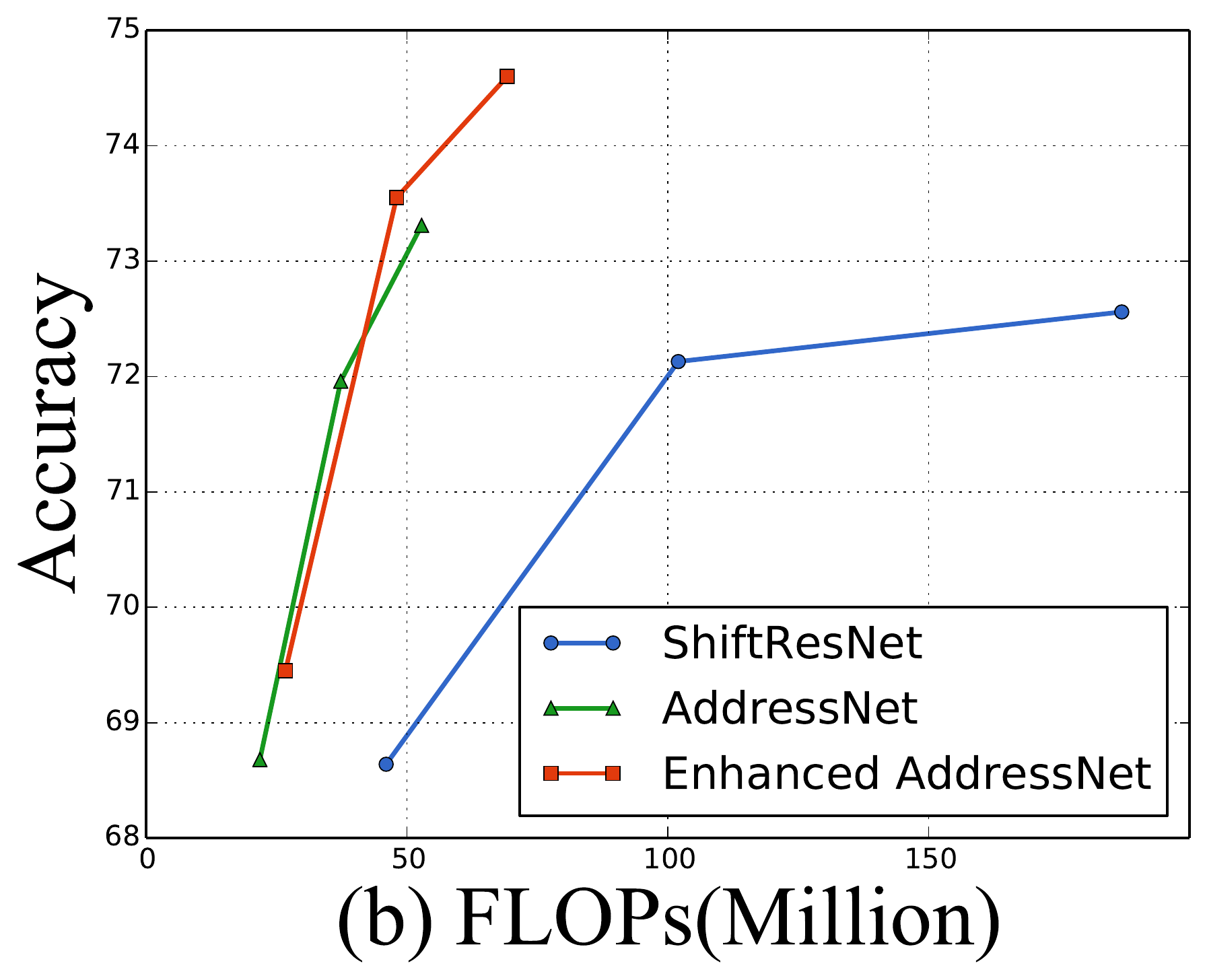}               
     \end{minipage}
   }
   \quad \quad
   \subfigure{
    \begin{minipage}{5cm}
     \centering    \includegraphics[scale=0.22]{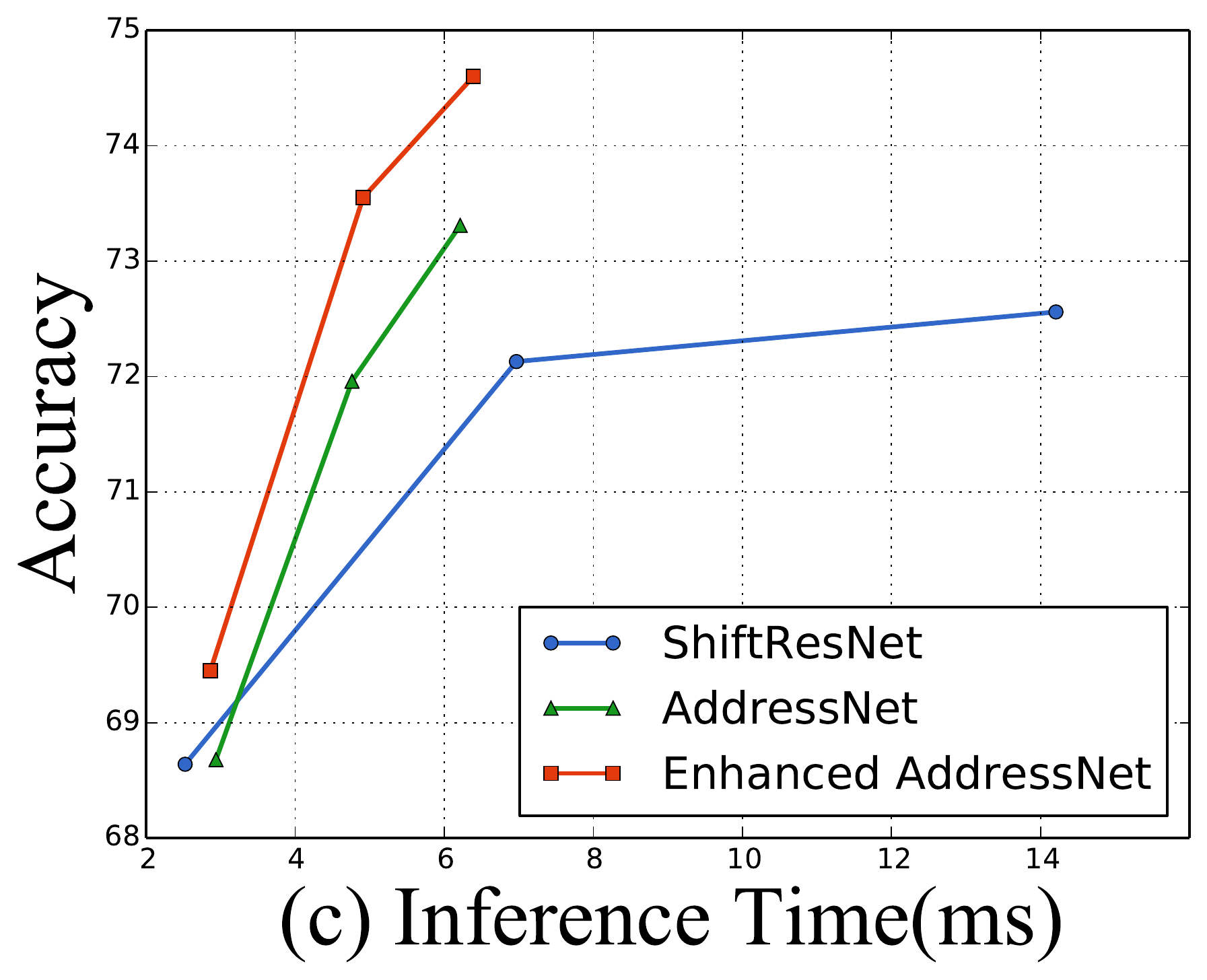} 
    \end{minipage}
   }
   
    \caption{A more clear comparison between \addressnet\  and ShiftResNet, summarized from Table~\ref{table:perf-shift-address}. These figures show that our network architectures are better than ShiftResNet family members with fewer parameters(a), FLOPs(b) and lower latency(c)} 
    \label{fig:compare-shift-address} 
 \end{figure*}

We evaluate \addressnet\ and \fastaddressnet\ with different depths on the CIFAR100 classification task, and compare it with ShiftResNet architecture. It is  reported in \cite{shiftnet} that ShiftResNet achieves better accuracy than ResNet  with similar architectures, but ShiftResNet requires fewer FLOPs and parameters. We elaborate three variants of \addressnet\ and \fastaddressnet\ on three different parameter scales and quote the results of ShiftResNet from \cite{shiftnet}. The results are shown in Table~\ref{table:perf-shift-address} and visualized in Fig~\ref{fig:compare-shift-address}. Compared to the best accuracy of ShiftResNet, our model(\addressnet-44) can achieve better performance with 3x fewer FLOPs and 6x fewer parameters. Furthermore, the curves in Fig~\ref{fig:compare-shift-address}(a) and Fig~\ref{fig:compare-shift-address}(b) show that our network architectures consistently obtain better accuracy than ShiftResNet in different parameters and FLOPs. In Fig~\ref{fig:compare-shift-address}(c), it presents that our models can reduce inference time significantly. Also note that the \fastaddressnet\ is always better than \addressnet~due to its larger groups and well-designed mechanism for shift operation as described in Section~\ref{s:architeture}.

\begin{table*}[t]
\begin{center}
\begin{tabular}{cccccc}
\hline\noalign{\smallskip}
Model & Parameters & FLOPs & Top1 Accuracy & GPU Time (ms)& CPU Time (ms)\\
\noalign{\smallskip}
\hline
\noalign{\smallskip}
ShiftResNet-20 & 0.19M & 46.0M & 68.64\% & 2.5$\pm$0.01 & 45.86$\pm$0.26\\
ShiftResNet-56 & 0.58M & 102M & 72.13\% & 7.0$\pm$0.04 & 115.45$\pm$1.50\\
ShiftResNet-110 & 1.18M & 187M & 72.56\% & 14.2$\pm$0.02 & 239.83$\pm$4.83\\
\hline
\addressnet-20 & 0.08M & 21.8M & 68.68\% & 2.9$\pm$0.01 & 10.73$\pm$0.01\\
\addressnet-32 & 0.14M & 37.3M & 71.96\% & 4.8$\pm$0.04 & 17.44$\pm$0.00\\
\addressnet-44 & {0.20M} & {52.8M} & {73.31\%} & {6.2}$\pm$0.08 &24.08$\pm$0.05\\
\hline
\fastaddressnet-20 & 0.18M & 26.7M & 69.45\% & 2.9$\pm$0.03 &16.86$\pm$0.66\\
\fastaddressnet-32 & 0.33M & 48.0M & 73.55\%  & 4.9$\pm$0.01 &25.90$\pm$0.01\\
\fastaddressnet-44 & 0.47M & 69.2M & {74.6\%} & {6.4}$\pm$0.01 &29.14$\pm$0.06\\
\hline
\end{tabular}
\end{center}
\caption{The performance of ShiftResNet and \addressnet\ in CIFAR100. The results of ShiftResNet are quoted from \cite{shiftnet}}
\label{table:perf-shift-address}
\end{table*}

\begin{table}[t]
\begin{center}
		\begin{tabu} to \hsize {c|c|c|c|c}
			\hline
			Type & Output size & Stride & $\varepsilon$ & Repeat \\ \hline \hline
			Input      & 224$\times$224, 3 & - & - & - \\ \hline
			Conv1      & 112$\times$112, 32 & 2 & - & 1\\ \hline
			Stage1     & 56$\times$56, 96   & 2 & 4 & 1 \\
			& 56$\times$56, 96   & 1 & 3 & 3 \\ \hline
			Stage2     & 28$\times$28, 192   & 2 & 3 & 1\\
			& 28$\times$28, 192   & 1 & 2 & 4 \\ \hline
			Stage3     & 14$\times$14, 384   & 2 & 2 & 1 \\
			& 14$\times$14, 384   & 1 & 2 & 5 \\ \hline
			Stage4     & 7$\times$7, 768     & 2 & 2 & 1 \\ 
			& 7$\times$7, 768     & 1 & 2 & 3 \\ \hline
			Pool & 1$\times$1, 768 & - & - & 1 \\ \hline
			FC & 1000 & - & - & 1 \\ \hline
		\end{tabu}
\end{center}
\caption{\fastaddressnet\ Architecture}
\label{table:addressnetArch}
\end{table}

\begin{figure*}[t] 
	\subfigure{  
		\begin{minipage}{5cm}
			\centering   
			\includegraphics[scale=0.22]{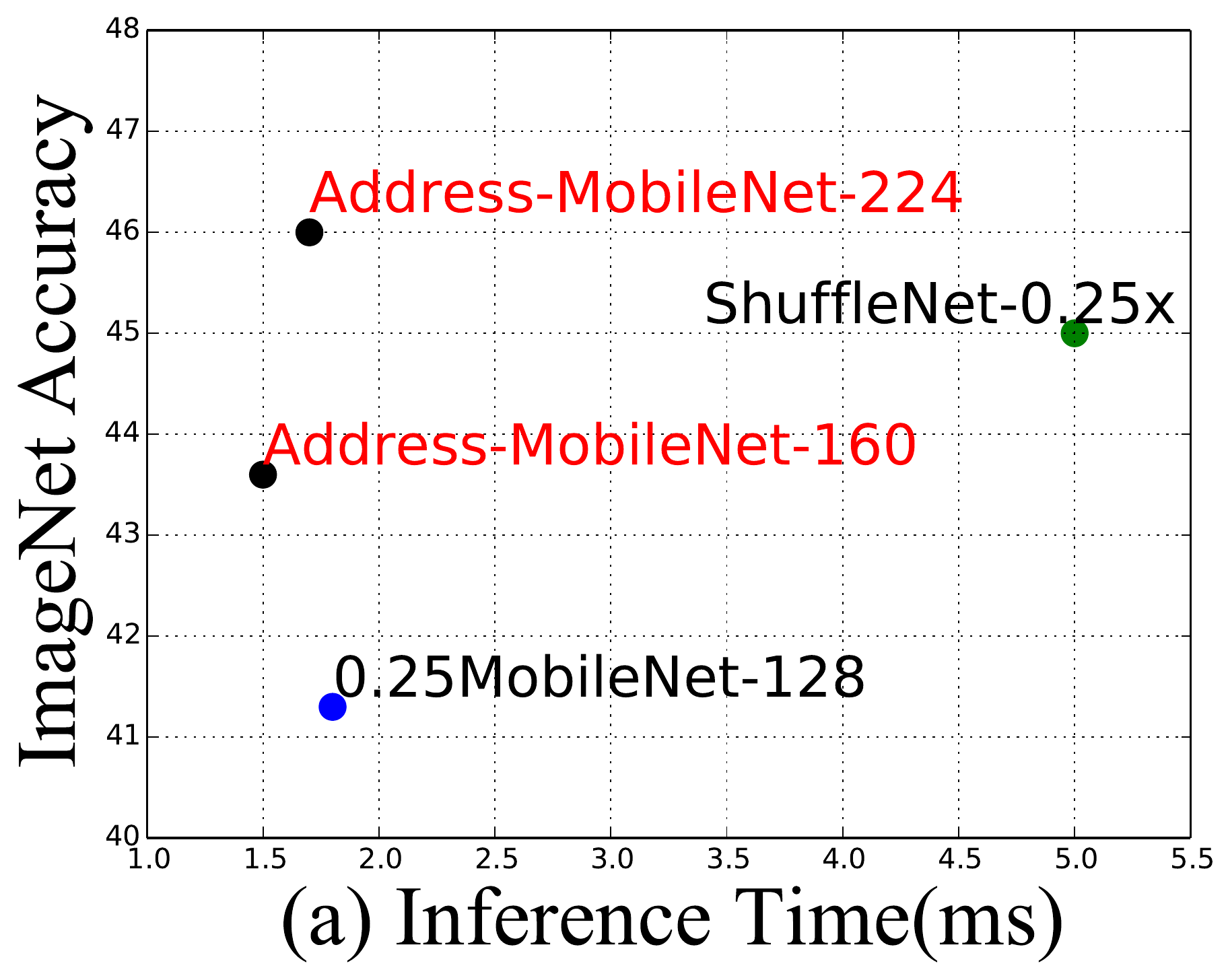}               
		\end{minipage}
	}
	\quad \quad
	\subfigure{  
		\begin{minipage}{5cm}
			\centering    
			\includegraphics[scale=0.22]{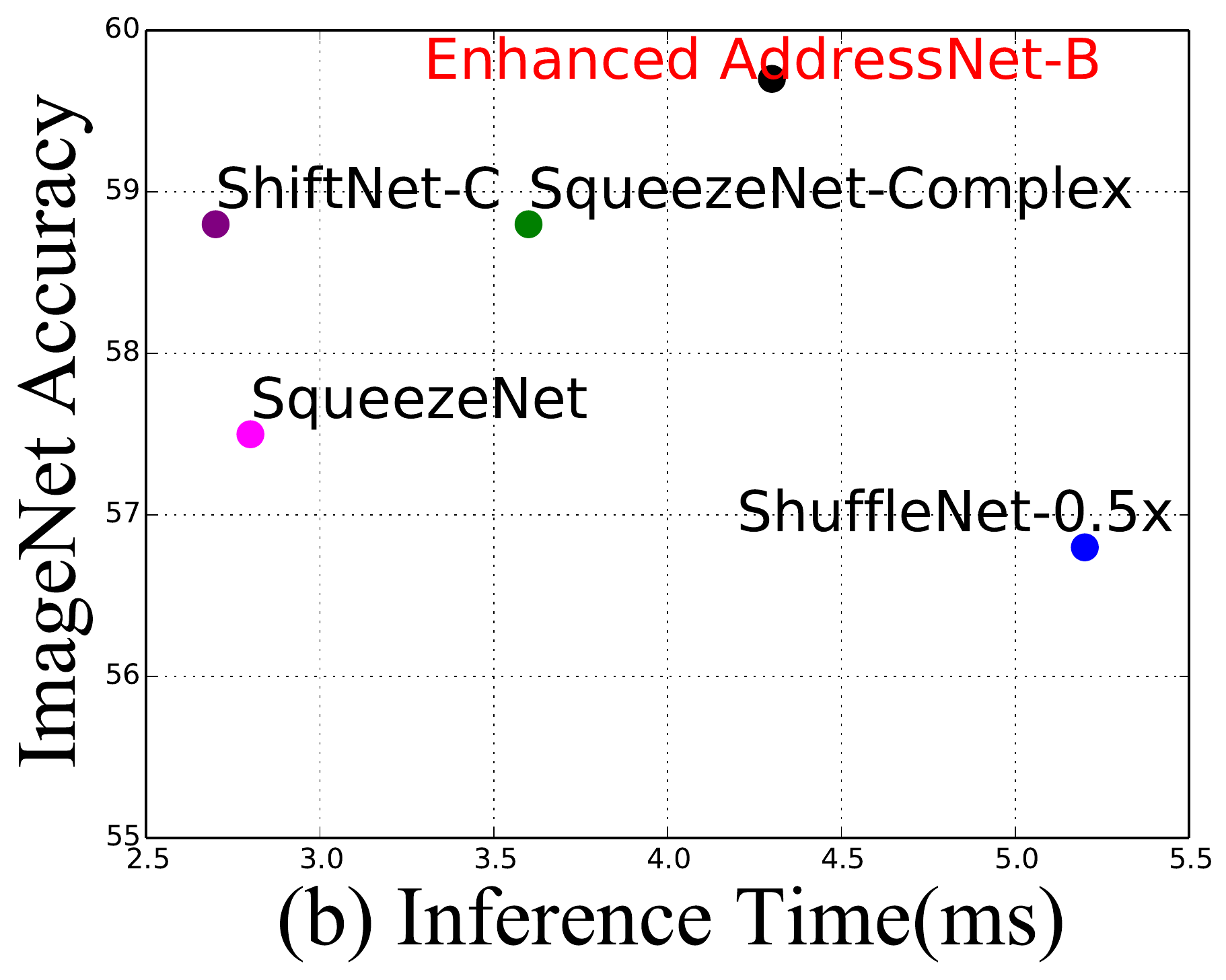}               
		\end{minipage}
	}
	\quad \quad
	\subfigure{
		\begin{minipage}{5cm}
			\centering    \includegraphics[scale=0.22]{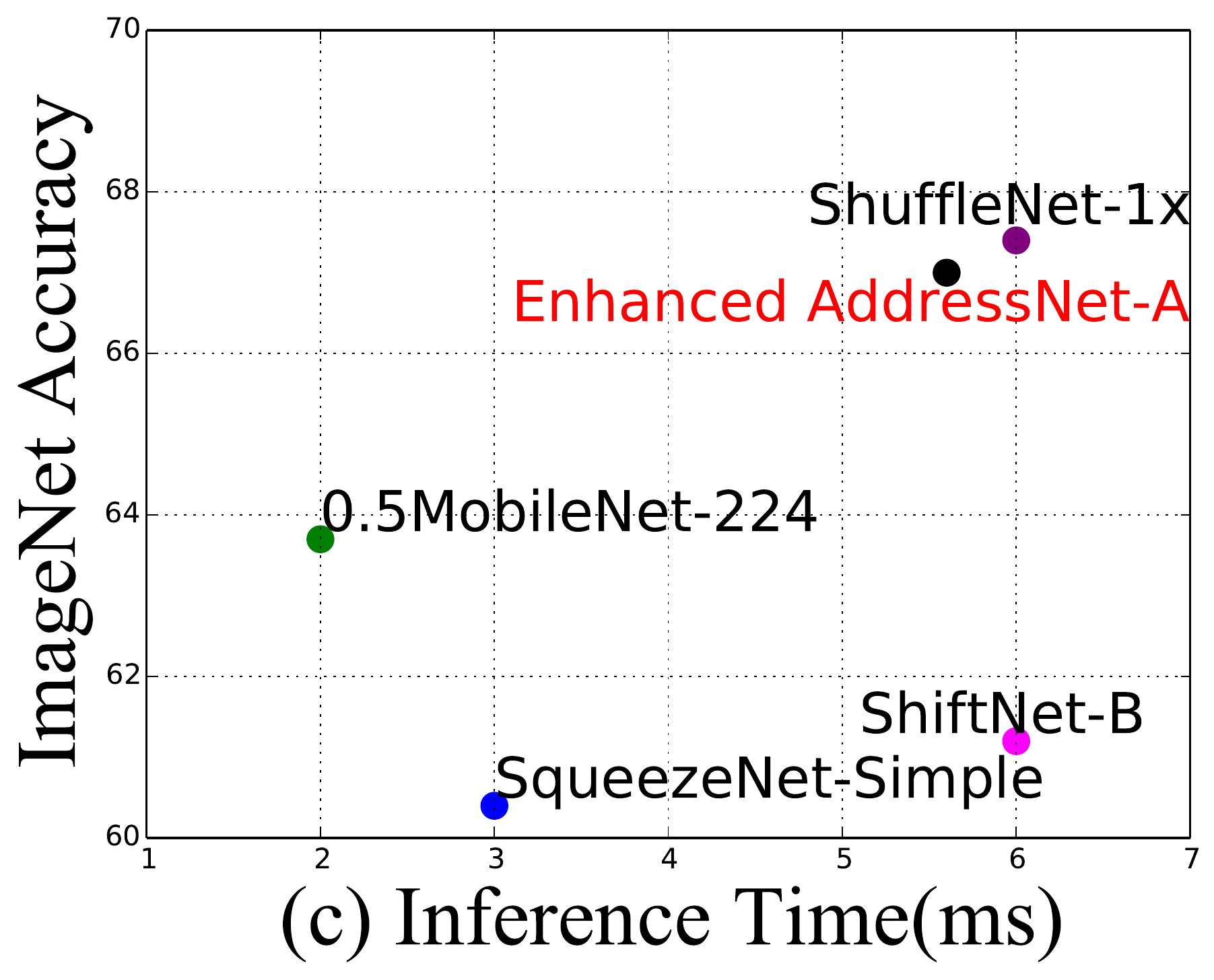} 
		\end{minipage}
	}
	
	\caption{The performance of different models in three levels of accuracy, under the reference frame of Inference Time and Accuracy. In this coordinate system, the closer to the left(time) and top(accuracy), the better is the model} 
	\label{fig:imagenet} 
\end{figure*}

\subsection{ImageNet}
Based on the above experiments, we have confirmed that our networks outperform ShiftNet~\cite{shiftnet} on CIFAR100 and the three shift operations can reduce parameters while retaining accuracy. To further assess the scalability and flexibility of our operations, we use our address shift operation to improve MobileNet and show its performance on ImageNet dataset. In our experiments on small models, we modify MobileNet to create \addmobile\ by doubling the output channel of the first convolution, removing the last $2\sim4$ layers, and replacing depthwise separable convolutions with address shift. Then similarly to MobileNet, we scale the input size to build \addmobile-192 and \addmobile-160. To adapt to ImageNet, we improve the depth and expansion rate  $\varepsilon$ of \fastaddressnet\ to form \fastaddressnet-A and \fastaddressnet-B. The detailed architecture of \fastaddressnet-A is listed in Table~\ref{table:addressnetArch} while  \fastaddressnet-B is a little shallower to fit the level of low accuracy and we will not show its details for simplicity.

The results for four levels of accuracy are shown in the Table~\ref{table:perf-imagenet} and their corresponding scatter diagram is also shown in Figure~\ref{fig:imagenet}. In the low-accuarcy scenario,  MobileNet, improved by our address shift operation, excels in both accuracy and inference time significantly (shown in Figure~\ref{fig:imagenet}(a)). 
This validates the scalability of the address shift operation on large dataset. In  the modest accuracy scenarios, our models achieve comparable performance with other state-of-the-art models with small and compact network architecture. 

In the NVIDIA CUDA Deep Neural Network library (cuDNN\footnote{https://developer.nvidia.com/cudnn}), there is little optimization for group convolution; therefore, this favors models that are not based on group convolution. 
That is why our models only achieve comparable results rather than the best results.  Generally speaking, our experiments on CIFAR100 and ImageNet demonstrates the value of using address shift to accelerate the inference process and also demonstrates our three shift operations have scalability and flexibility for designing compact architectures.

\begin{table}[t]
\begin{center}
\begin{tabular}{ccc}
\hline\noalign{\smallskip}
Model & \begin{tabular}[c]{@{}c@{}}Top-1\\ Acc.\end{tabular} & \begin{tabular}[c]{@{}c@{}}Latency\\ (ms)\end{tabular}\\
\noalign{\smallskip}
\hline
\noalign{\smallskip}
ShuffleNet-1x & \textbf{67.4}\% & 6.0 \\
\textbf{\fastaddressnet-A(Ours)} & 67.0\% & 5.6 \\
0.5MobileNet-224 & 63.7\%& \textbf{2.0}\\
ShiftNet-B & 61.2\% & 6.0\\
SqueezeNet-Simple & 60.4\% & 3.0\\

\hline
\textbf{\fastaddressnet-B(Ours)} & \textbf{59.7}\% & 4.3\\
ShiftNet-C & 58.8\% & \textbf{2.7}\\
SqueezeNet-Complex & 58.8\% & 3.6\\
SqueezeNet & 57.5\% & 2.8\\
ShuffleNet-0.5x & 56.8\% & 5.2 \\
\hline
\textbf{\addmobile-224(Ours)} & \textbf{46.0}\% & 1.7 \\
ShuffleNet-0.25x & 45.0\% & 5.0 \\
\textbf{\addmobile-160(Ours)} & 43.6\% & \textbf{1.5} \\
0.25MobileNet-128 & 41.3\%& 1.8\\
\hline
\end{tabular}
\end{center}
\caption{The performance of different levels of accuracy on ImageNet. We modify MobileNet with address shift operation to build \addmobile\ family and improve  \fastaddressnet\ to build \fastaddressnet-A and \fastaddressnet-B. \fastaddressnet-B is a little shallower to fit the level of low accuracy}
\label{table:perf-imagenet}
\end{table}

\section{Conclusions}
The practical value of using deep neural networks in latency-sensitive power and energy constrained embedded systems naturally motivates the search for techniques to create fast, energy-efficient deep neural-net models. This paper adds a collection of shift based operations, namely, channel shift, shortcut shift and address shift, to the set of useful techniques for designing such models. In particular, these shift-based techniques utilize  address offset to realize spatial convolutions and  no parameters and no FLOPs, and thereby no inference time. Based on these shift operations, we proposed two inference-efficient CNN models named \addressnet\ and \fastaddressnet, which outperform ShiftNet significantly. We have also used our operations to improve state of the art in neural net architecture: on CIFAR and ImageNet datasets we demonstrate that, deep neural nets designed with our shift operations can achieve better accuracy. In the future, we plan to develop a library to optimize group convolution which will make it more adaptable to our shift operations. 

{\small
\bibliographystyle{ieee}
\bibliography{egbib}
}

\end{document}